\documentclass[twoside,11pt]{article}

\usepackage{automl2020}
\usepackage{custom}

\jmlrheading{\shortauthors\ }

\ShortHeadings{Automated LR Scheduler for Large-batch Training}{\shortauthors\ }
\firstpageno{1}

\begin{document}

\title{Automated Learning Rate Scheduler for Large-batch Training}

\authors

\maketitle

\begin{abstract}
Large-batch training has been essential in leveraging large-scale datasets and models in deep learning. While it is computationally beneficial to use large batch sizes, it often requires a specially designed learning rate (LR) schedule to achieve a comparable level of performance as in smaller batch training. Especially, when the number of training epochs is constrained, the use of a large LR and a warmup strategy is critical in the final performance of large-batch training due to the reduced number of updating steps. 
In this work, we propose an automated LR scheduling algorithm which is effective for neural network training with a large batch size under the given epoch budget. In specific, the whole schedule consists of two phases: adaptive warmup and predefined decay, where the LR is increased until the training loss no longer decreases and decreased to zero until the end of training. Here, whether the training loss has reached the minimum value is robustly checked with Gaussian process smoothing in an online manner with a low computational burden.
Coupled with adaptive stochastic optimizers such as AdamP and LAMB, the proposed scheduler successfully adjusts the LRs without cumbersome hyperparameter tuning and achieves comparable or better performances than tuned baselines on various image classification benchmarks and architectures with a wide range of batch sizes.
\end{abstract}

\section{Introduction}

In modern deep learning tasks, scaling up both the sizes of dataset and model has shown promising improvements. However, a large number of samples and slow gradient computations lead to considerably longer training time, and therefore the use of large batch size in stochastic optimization with multiple computational nodes has gain popularity to speed up for such a large-scale training. It is desirable for the optimization algorithm with the increased batch sizes to maintain the performances without increasing the amount of processed samples, namely the number of training epochs.

Recently, several works have been proposed for successful large-batch training with stochastic gradient descent (SGD), and most of them achieve it by specially designed learning rate (LR) schedules and especially LR scaling \citep{krizhevsky2014weird, goyal2017gradual, hoffer2017longer}.
For instance, linear~\citep{krizhevsky2014weird, goyal2017gradual} or square-root~\citep{hoffer2017longer} LR scaling rules according to the batch sizes effectively alleviate the performance degradation in large-batch training by compensating the reduced number of optimization steps. In addition, gradual LR warmup heuristic \citep{goyal2017gradual} reduces the instability caused by large LRs and has become a standard for large-batch training. More recently, layer-wise adaptive LR scalings have been proposed to further increase the batch size and are applied to diverse tasks \citep{you2017lars, you2020lamb, ginsburg2019novograd}. However, these LR scaling and warmup heuristics are sensitive to its hyperparameters including the LR and the warmup schedule, hence require intensive tuning effort.

There are a number of studies attempting to automate the LR scheduling to reduce such cumbersome tuning and to enhance the performance. Online or offline LR search algorithms using Bayesian optimization \citep{victor2020bo} or meta-optimization \citep{schraudolph1999local, chen2017l2l, gunes2018hd, donini2020marthe} have been proposed. However, they have been limited to small (surrogate) tasks due to complexity and the feasibility of them in large-batch training is unknown. 

We propose an automated online LR scheduler for large-batch training. The schedule consists of the warmup phase, where the LR is increased until the training loss no longer decreases, and the decay phase that decreases the LR. To robustly decide the phase transition on the fly, we employ Gaussian process (GP) smoothing \citep{rasmussen2006gaussian}. This GP-based online detection is robust to small bumps and has a low computational burden. 
In addition, to cover a wide range of possible LR values while ensuring stability, the LR is exponentially increased from a very small value ($10^{-5}$) to a very large value (1.0) up to the half of the given epoch budget. As a result, the proposed LR scheduler can efficiently and automatically figure out not only the initial and peak LRs but also the warmup length in a data-driven way as the training progresses. From the perspective of having the warmup and the decay phase, SALSA \citep{zhang2020salsa} is most similar to this work. However, SALSA relies on the backtracking line search for the warmup phase which can be burdensome, and it was not evaluated on large batch sizes. 

We empirically demonstrate that the proposed LR scheduler together with an adaptive optimizer such as AdamP \citep{heo2021adamp} or LAMB \citep{you2020lamb} achieves comparable or better performance compared to a tuned baseline in case of large batch size on various image classification benchmarks and architectures.

\vspace{-0.3cm}
\section{Methods}

\begin{figure}[t]
\centering
\includegraphics[scale=0.5]{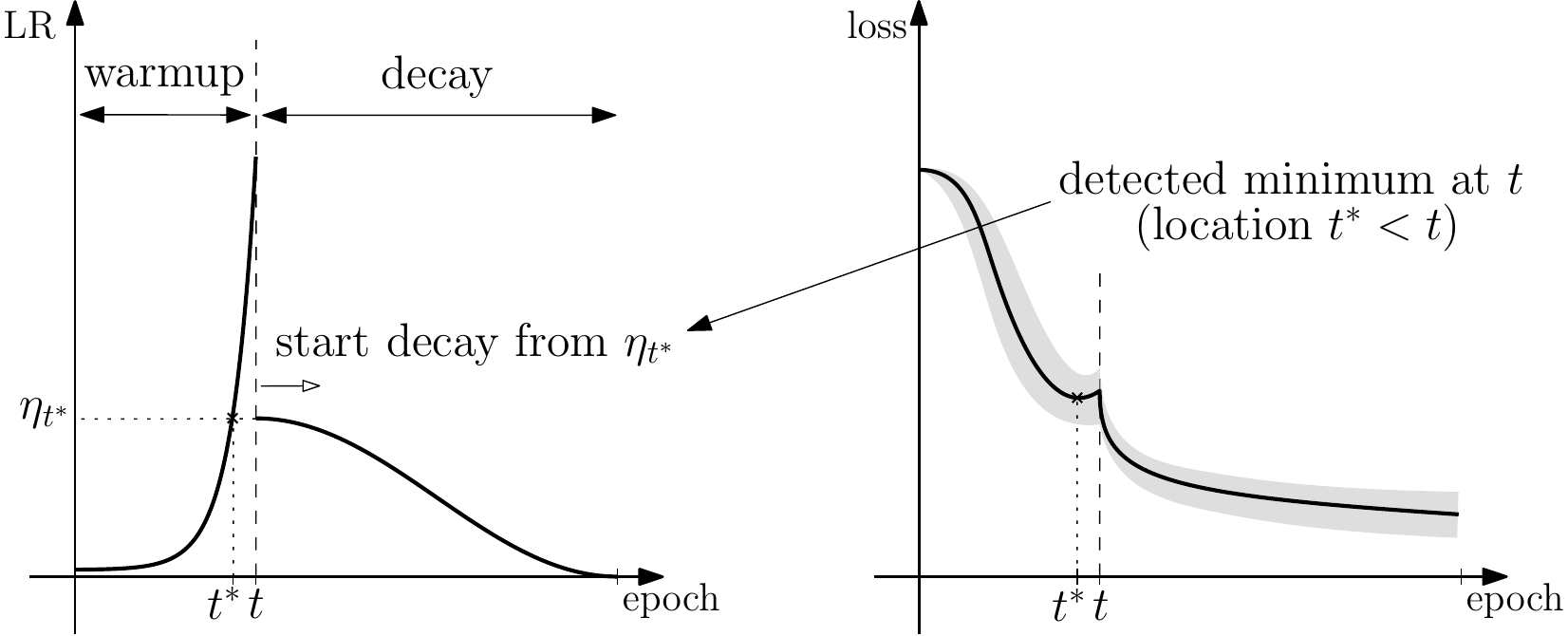}
\vspace{-0.2cm}
\caption{Conceptual diagram of \autowu. It adaptively switches from the warmup phase to a decay schedule (\emph{e.g.} cosine) at $t$, upon detection of minimum loss at $t^*$ preceding $t$ with high probability. The start LR of the decay phase is also adaptively set as $\eta_{t^*}$ corresponding to $t^*$. The shaded region in the right plot indicates the variance of the observed losses, and the bold curve corresponds to the smoothed loss via GP.}
\label{fig:concept}
\vspace{-0.3cm}
\end{figure}

We propose an automated LR scheduler, called \autowu\ (stands for automated warmup), which consists of two phases \emph{warmup} and \emph{decay}.
Figure \ref{fig:concept} describes how \autowu\ works.
\vspace{-0.2cm}

\subsection{Warmup Phase}
\label{subsec:methods:warmup-phase}
During the warmup phase, the LR starts from a very small value and is increased following an exponential schedule. It is tested at the end of each epoch whether the loss has hit the minimum value as training progresses. If the minimum has been detected, then the LR follows the predefined decay schedule. This scheme combines the warmup strategy which enhances the stability in the early phase of training with the automated LR selection.

\subsubsection{Warmup by exponential schedule}
The proposed exponential schedule in the warmup phase has three hyperparameters: the initial LR $\eta_{\min}$, the maximum possible LR $\eta_{\max}$, and the maximum possible fraction $\rho_w \in (0, 1)$ of warmup steps. Given the total number of steps $T$, the warmup schedule is defined as $\eta_0 = \eta_{\min}$ and
\begin{equation}
\label{eqn:warmup_sched}
    \eta_t = \gamma \eta_{t-1} 
    \quad \text{for} \quad 
    t \in \{1,...,\lfloor \rho_w T \rfloor\},
    \quad \text{where}\quad
    \gamma = \left(\eta_{\max} / \eta_{\min} \right)^{1/\lfloor \rho_w T\rfloor}.
\end{equation}
Here, if the training loss keeps decreasing, then the warmup phase can last for $\lfloor \rho_w T \rfloor$ steps, and $\eta_t$ can grow up to $\eta_{\max}$. 

In comparison to the other growth rates such as the linear growth rate, this exponential growth rate enables stable and fine-grained LR exploration, especially in the early stage for large-batch training. We set $\eta_{\min} = 10^{-5}$ and $\eta_{\max} = 1$ to ensure that LRs can sweep a wide enough range of values.\footnote{This range of LRs is sufficient for AdamP and LAMB, but may require a modification for other optimizers, e.g., a good value of $\eta_{\max}$ for SGD would be larger.}
Moreover, we set $\rho_w = 0.5$ to ensure that the LR does not grow too fast. Results regarding the sensitivity of \autowu\ with respect to $\rho_w$ and the comparison with the linear growth are found in Appendix \ref{app:abl_warmup}.

\subsubsection{Online minimum detection via Gaussian process}
\label{subsec:methods:gp}

A loss trajectory may exhibit local and global fluctuations due to the stochastic nature of a mini-batch computation and a highly non-convex loss landscape. This makes it difficult to robustly detect whether a loss is no longer decreasing or not in an online manner. As a remedy, we propose to use GP regression to smooth the loss curve and then conduct a decision test based on the predictive distribution.

Suppose that we have been observed (noisy) loss values $L_0, \cdots, L_{t-1}$ at step $t$. We model the loss curve $s \mapsto L_s$ (defined for $s \in [0, t]$) by a GP with a homoskedastic noise:
\begin{equation}
    L_s = f(s/t) + \epsilon ,
    \quad \text{where} \quad
    f \sim \mathcal{GP}(\theta, K_{\ell, \sigma_f}(\cdot, \cdot))
    \text{ and }
    \epsilon \sim \mathcal{N}(0, \sigma_n^2),
\end{equation}
where $\theta$, $K_{\ell, \sigma_f}$ and $\sigma_n^2$ denote the mean, the covariance kernel, and the noise level, respectively. When conditioned on the observed sequence $L_0, \cdots, L_{t-1}$, the predictive distribution of $f$ is again a GP and can be computed straightforwardly.\footnote{See \citet{rasmussen2006gaussian} for a good exposition on the subject.}

We are interested in whether $f$ is no longer decreasing, \emph{i.e.}, there exists $x^* < 1$ such that $f(x^*) < f(1)$. Hence, it is natural to compute the probability
\begin{equation}
\label{eqn:true_prob}
    \mathbb{P}_f\left( f(x) < f(1) \text{ for some }x \in [0, 1]\right),
\end{equation}
and decide to end the warmup phase if (\ref{eqn:true_prob}) is large. However, since (\ref{eqn:true_prob}) is not easy to compute, we compute the following lower bound instead:
\begin{equation}
\label{eqn:lb_prob}
    P_{\min}(f) := \max_{x \in S} \mathbb{P}_f(f(x) < f(1)),
\end{equation}
which holds for any $S \subseteq [0, 1]$. Here, $S$ is chosen to be 500 equally-spaced points in $[0, 1]$.

\paragraph{GP model details.}
We use the GP model with the constant mean $\theta$ and the squared-exponential kernel $K_{\ell, \sigma_f}(x, x') = \sigma_f^2 \cdot e^{-|x-x'|^2 / (2\ell^2)}$, where the length-scale $\ell$ is fixed to $0.2$ to prevent model overfitting to local variations, and all other parameters $\theta$, $\sigma_f$, and $\sigma_n$ are fitted by Adam \citep{kingma2014adam} with LR $0.01$ for 100 steps.

\paragraph{Test details.} 
There are three main hyperparameters involved in the test: $n_{test}$, confidence $c$, and patience $p$. We used $n_{test}=5$, $c=0.95$ and $p=3$ in all experiments.

Suppose that the loss trajectory $\mathcal{C} = \{(0, L_0), \cdots, (t-1, L_{t-1})\}$ has been observed at step $t$. To conduct the test, we first choose a random subset $\mathcal{C}_0 \subseteq \mathcal{C}$ of maximum 100 samples and fit the GP parameters with respect to the marginal log-likelihood of $\mathcal{C}_0$. Then, we further sample $n_{test}$ random subsets $\mathcal{C}_1, \cdots, \mathcal{C}_{n_{test}}$ of $\mathcal{C}$, each of maximum 500 samples, and infer $f_i$ conditioned on $\mathcal{C}_i$ for $i=1,\cdots,n_{test}$. Finally, we compute $P_{\min}(f_i)$ in Eqn. (\ref{eqn:lb_prob}) and if the majority of $P_{\min}(f_i)$'s exceed the confidence $c$, then it is regarded as that the minimum has been detected.

We start the decay phase only if the minimum has been detected in $p$ consecutive tests. This is to prevent premature ending of warmup due to random spikes in the loss trajectory, which was observed more often in large-batch setting in our preliminary experiments. We remark that larger $p$ implies the switch happening later; this may result in a too large LR. To remove the impact of the choice of $p$, we set the starting LR of the decay phase to be $\eta_{t^*} = \gamma^{t^*} \eta_0$ where
\begin{equation}
\label{eqn:tstar}
    t^* := t \cdot \mathbb{E}_{i} \left[ \arg \min_{x \in [0, 1]} \mathbb{E}_{f_i}\left[ f_i(x)\right] \right].
\end{equation}
The overall algorithm is summarized in Appendix \ref{app:details} (Algorithm \ref{alg:main}).

\paragraph{Computation time.}
Time complexity of exact GP inference increases cubically with the number of samples. Nonetheless, those computations are not very burdensome since they are only conducted once per epoch in this work, and we use subsampling so that the number of samples is kept constant regardless of the task (particularly the batch size). Moreover, we use GPyTorch \citep{gardner2018gpytorch} for efficient GP computations on GPUs. Time overhead per test (fitting and 5 inferences) was less than a second in average, which is typically much smaller than the overall gradient computation time per epoch.
\vspace{-0.2cm}
\subsection{Decay Phase}
\label{subsec:methods:decay-phase}
In the decay phase, \autowu\ follows an LR schedule with the predefined shape, but whose starting LR is adaptively determined in the warmup phase. To remove any sophistication in evaluation, we only consider two simple types of schedule in the decay phase: \emph{cosine} \citep{loshchilov2016sgdr} or \emph{constant-then-cosine}.
In case of cosine decay, the LR starts with the value determined by the warmup phase and is annealed toward zero with cosine schedule. This schedule is particularly appealing since it does not introduce any additional hyperparameters to consider. 
On the other hand, constant-then-cosine decay maintains LR constant until a predetermined fraction of epochs is left and then follows cosine schedule. We set the fraction 0.2 (\emph{i.e.} cosine decay in the last 20\% of epochs) in all experiments. We have empirically observed that an automated LR decay based on the convergence test such as SASA+ \citep{zhang2020salsa} was not decisively superior than these schedules. This probably implies that a warmup strategy including the peak LR is more critical for the final performance, especially in large-batch training setting.

\vspace{-0.2cm}
\section{Experiments}
\label{sec:experiments}

We evaluate our algorithm on image classification tasks with three benchmark datasets: CIFAR-10, CIFAR-100 \citep{krizhevsky2009learning} and ImageNet \citep{deng2009imagenet}. We mainly consider convolutional networks: ResNet-18, Wide-ResNet-28-10 \citep{zagoruyko2016wide} for CIFAR-10 and CIFAR-100 respectively, and ResNet-50 \citep{he2016deep} for ImageNet. We also conduct evaluations on ImageNet with EfficientNet-B0 \citep{tan2019efficientnet} and a vision transformer ViT-S/16 \citep{dosovitskiy2020image}, and the results by these other architectures are included in Appendix \ref{app:further_eval}.

\autowu\ only requires to know how the loss values are changing over the course of training, hence it is capable to be used with any stochastic descent algorithm. In our evaluations, the state-of-the-art Adam-based optimizer, AdamP \citep{heo2021adamp} is used as the base optimizer due to their better stability than SGD and its variants with large batch sizes.\footnote{In our preliminary experiments, we have observed that instability of SGD could be mitigated with gradient clipping but extensive tuning effort was required to match the same level of performances.} The results when coupled with the layer-wise adaptive optimizer, LAMB \citep{you2020lamb}, are also presented in Appendix \ref{app:further_eval}.

\begin{table*}[t]
\scriptsize
    \centering
    \caption{Comparison of test/val accuracies (\%) between the baseline schedule and \autowu\ with two decay schedules (cosine and constant-then-cosine) when used with AdamP. We report the mean and the standard deviation (written in parenthesis) of three independent runs with different random seeds in case of CIFAR tasks.}
    \begin{tabular}{cc|cccc}
    \hline
    \multirow{2}{*}{\makecell{Dataset \\ (Architecture)}}
    & \multirow{2}{*}{Schedule} 
    & \multicolumn{4}{c}{Batch size} 
    \\
    & & 256 & 1K & 8K & 16K
    \\
    \hline
    \multirow{3}{*}{\makecell{CIFAR-10 \\ (ResNet-18)}} 
    & Baseline 
    & \textbf{96.58 {\tiny (0.07)}} 
    & \textbf{96.48 {\tiny (0.02)}} 
    & \textbf{96.05 {\tiny (0.15)}}
    & 94.63 {\tiny (0.06)}
    \\
    & \autowu\ + const-cos
    & 96.26 {\tiny (0.12)}
    & 96.20 {\tiny (0.03)}
    & 95.92 {\tiny (0.22)}
    & \textbf{94.80 {\tiny (0.17)}}
    \\
    & \autowu\ + cos
    & 96.43 {\tiny (0.02)}
    & 96.42 {\tiny (0.05)}
    & 95.77 {\tiny (0.01)}
    & 94.03 {\tiny (0.26)}
    \\
    \hline
    \multirow{3}{*}{\makecell{CIFAR-100 \\ (Wide-ResNet28-10)}} 
    & Baseline
    & 83.36 {\tiny (0.38)}
    & 83.13 {\tiny (0.14)}
    & 81.08 {\tiny (0.33)}
    & 77.62 {\tiny (0.36)}
    \\
    & \autowu\ + const-cos
    & 83.36 {\tiny (0.21)}
    & 83.21 {\tiny (0.19)} 
    & \textbf{82.32 {\tiny (0.42)}}
    & \textbf{81.42 {\tiny (0.35)}}
    \\
    & \autowu\ + cos
    & \textbf{83.59 {\tiny (0.46)}}
    & \textbf{83.39 {\tiny (0.20)}} 
    & 82.26 {\tiny (0.60)}
    & 80.25 {\tiny (0.36)}
    \\
    \hline
    & & 1K & 4K & 16K & 32K \\
    \hline
    \multirow{3}{*}{\makecell{ImageNet \\ (ResNet-50)}}  & Baseline 
    & 76.28 & 76.10 & 75.02 & 74.11
    \\
    & \autowu\ + const-cos
    & \textbf{76.31} & \textbf{76.33} & \textbf{75.62} & \textbf{74.84}
    \\
    & \autowu\ + cos
    & 76.19 & 75.70 & 75.22 & 74.40
    \\
    \hline
    \end{tabular} 
    \label{tab:main_exp}
\normalsize
\vspace{-0.5cm}
\end{table*}

The proposed \autowu\ scheduler is compared with the conventional baseline LR schedule on each task with multiple batch sizes: $\{256, 1024, 8192, 16384\}$ for CIFAR (200 epochs) and $\{1024, 4096, 16384, 32768\}$ for ImageNet (120 epochs). 
The baseline LR schedule consists of 5 epochs of a linear warmup from 0 to a predetermined peak LR followed by a cosine decay to 0. Here, following a common LR scaling practice for large-batch training, the peak LR is scaled with the batch-size according to the square-root scaling rule, specifically, $\eta_{base} \sqrt{B/256}$ for batch size $B$. We set $\eta_{base}=0.001$ in all experiments by empirically tuning it as the common base LR that is well performed across tasks and architectures. We verify that the base LR as well as the warmup length strongly affect the performances of the baseline LR scheduler for large-batch training (see Appendix \ref{app:abl_grid_search}), however the proposed \autowu\ scheduler removes these tuning efforts.

As shown in Table \ref{tab:main_exp}, the proposed \autowu\ performs on par with or better than the baseline LR schedule across diverse batch sizes and tasks. Especially, with large batch sizes of 8K and 16K for CIFAR-100 and 16K and 32K for ImageNet, \autowu\ with constant-then-cosine decay significantly reduces the performance drops in comparison to the conventional baseline. In addition, overall, the constant-then-cosine decay is slightly better than the cosine decay when combined with \autowu\, which means that as we find the safe peak LR automatically for a long warmup time, retaining the found LR longer would be better.

\begin{figure}[t]
    \centering
    \begin{subfigure}[b]{0.3\textwidth}
    \includegraphics[width=\textwidth]{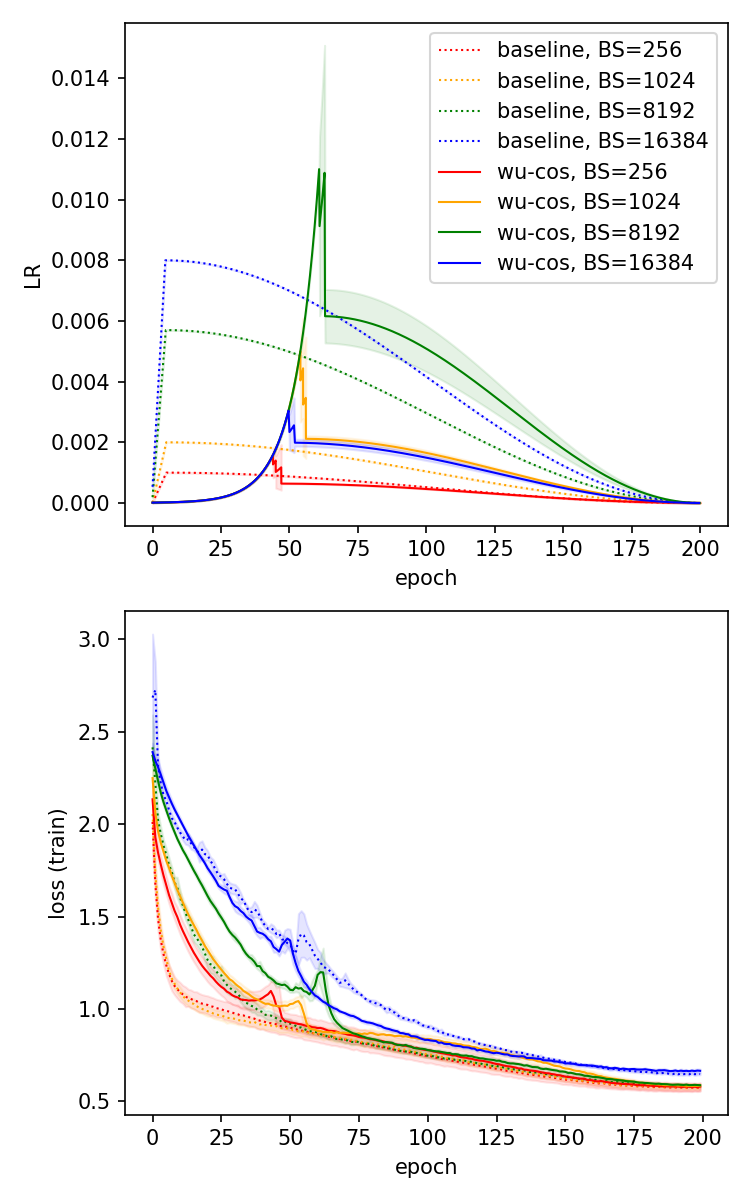}
    \caption{CIFAR-10}
    \end{subfigure}
    \hfill
    \begin{subfigure}[b]{0.3\textwidth}
    \includegraphics[width=\textwidth]{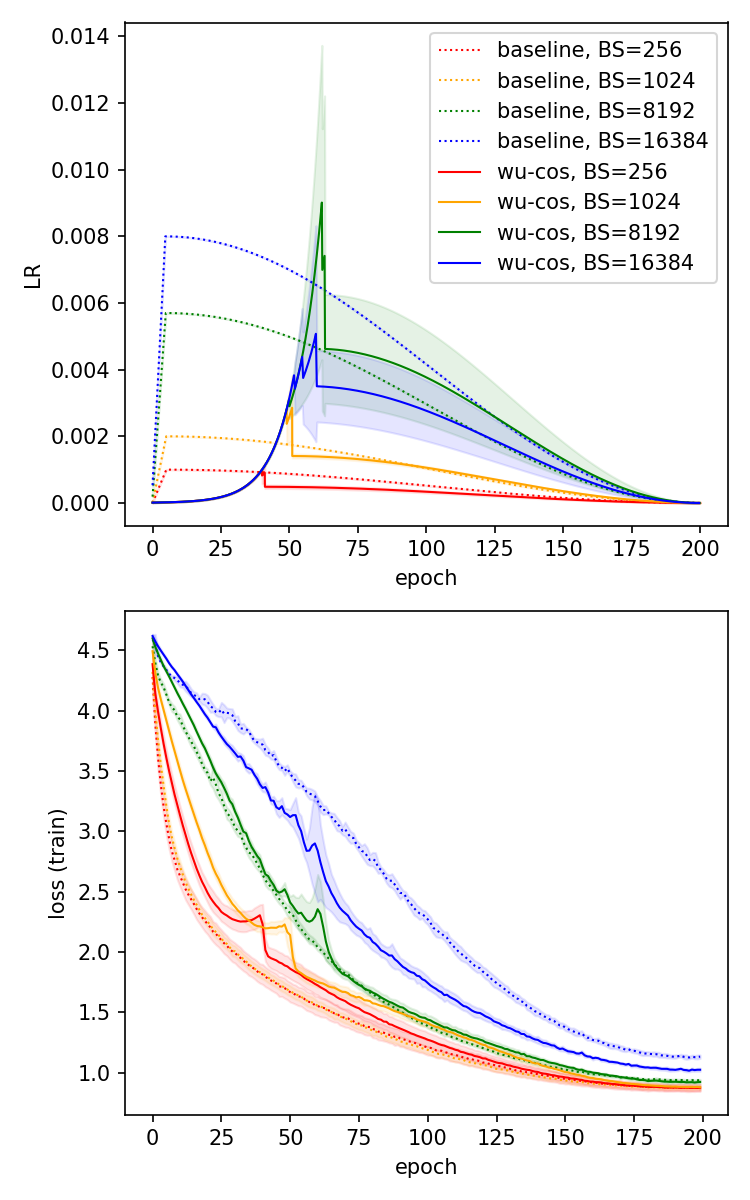}
    \caption{CIFAR-100}
    \end{subfigure}
    \hfill
    \begin{subfigure}[b]{0.3\textwidth}
    \includegraphics[width=\textwidth]{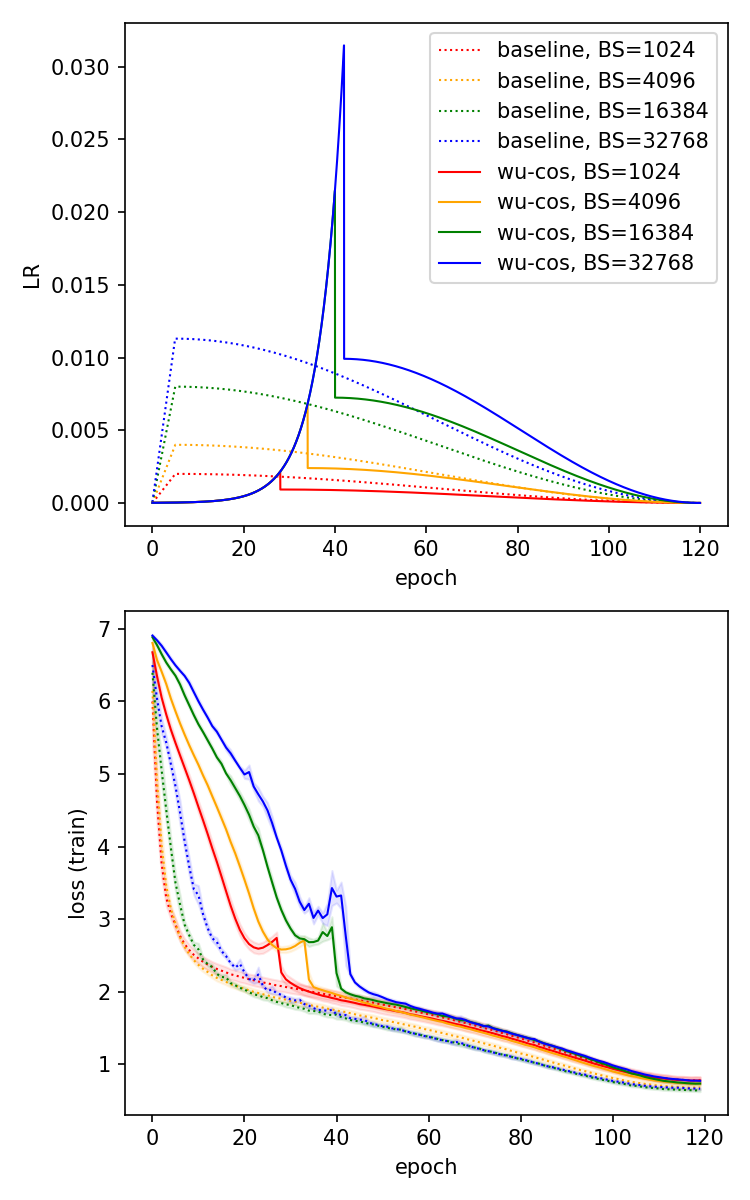}
    \caption{ImageNet}
    \end{subfigure}
    \caption{LR schedules (top row) and training loss curves (bottom row) of the baseline and \autowu\ with cosine decay (denoted wu-cos in the legend) on CIFAR and ImageNet tasks. Shadowed area represents the standard deviation computed over different seeds.
    }
    \label{fig:main_exp}
    \vspace{-0.5cm}
\end{figure}

Figure \ref{fig:main_exp} shows the LR schedule that \autowu\ found and the training loss curve in each task. For any batch size and task, \autowu\ robustly detects the minimum loss in an online manner by GP. Also, interestingly, the starting LR of the decay phase by \autowu\ grows as the batch size increases, and the rate is similar to the baseline scaling of the peak LR even though the warmup epochs are quite different. Actually, the found combination of the peak LR and the warmup length is similar to the search results on the baseline schedule as shown in Appendix \ref{app:abl_grid_search}. More ablations on \autowu\ are presented in Appendix \ref{app:abl_studies}. 
\vspace{-0.2cm}

\section{Conclusion}
In this work, an automated LR scheduler especially for fast training using large batch sizes is proposed. The proposed LR schedule has the warmup phase followed by the decay phase, where the LR is exponentially increased until the training loss no longer decreases and then decreased to zero until the end of training. The online detection of the minimum loss has been efficiently and robustly realized by GP regression. Empirical evaluation demonstrates that our automated LR scheduler appropriately adapts the whole warmup procedure as well as LRs for any batch size and task, and consequently results in comparable or better performances in comparison to the fine-tuned LR schedulers. Our implementation will be available at {\color{red} \url{https://github.com/kakaobrain/autowu}}.

\acks{We would like to acknowledge and thank Woonhyuk Baek and Brain Cloud Team at Kakao Brain for their support.}

\vskip 0.2in
\bibliography{mainbib}

\begin{thebibliography}{30}
\providecommand{\natexlab}[1]{#1}
\providecommand{\url}[1]{\texttt{#1}}
\expandafter\ifx\csname urlstyle\endcsname\relax
  \providecommand{\doi}[1]{doi: #1}\else
  \providecommand{\doi}{doi: \begingroup \urlstyle{rm}\Url}\fi

\bibitem[Baydin et~al.(2018)Baydin, Cornish, Rubio, Schmidt, and
  Wood]{gunes2018hd}
Atilim~Gunes Baydin, Robert Cornish, David~Martinez Rubio, Mark Schmidt, and
  Frank Wood.
\newblock Online learning rate adaptation with hypergradient descent.
\newblock In \emph{International Conference on Learning Representations}, 2018.
\newblock URL \url{https://openreview.net/forum?id=BkrsAzWAb}.

\bibitem[Chen et~al.(2021)Chen, Xie, and He]{xinlei20201mocov3}
Xinlei Chen, Saining Xie, and Kaiming He.
\newblock An empirical study of training self-supervised vision transformers.
\newblock \emph{arXiv preprint arXiv:2104.02057}, 2021.

\bibitem[Chen et~al.(2017)Chen, Hoffman, Colmenarejo, Denil, Lillicrap,
  Botvinick, and de~Freitas]{chen2017l2l}
Yutian Chen, Matthew~W. Hoffman, Sergio~G{\'o}mez Colmenarejo, Misha Denil,
  Timothy~P. Lillicrap, Matt Botvinick, and Nando de~Freitas.
\newblock Learning to learn without gradient descent by gradient descent.
\newblock In \emph{Proceedings of the 34th International Conference on Machine
  Learning}, volume~70 of \emph{Proceedings of Machine Learning Research},
  pages 748--756. PMLR, 2017.
\newblock URL \url{http://proceedings.mlr.press/v70/chen17e.html}.

\bibitem[Cohen et~al.(2021)Cohen, Kaur, Li, Kolter, and
  Talwalkar]{cohen2021gradient}
Jeremy Cohen, Simran Kaur, Yuanzhi Li, J~Zico Kolter, and Ameet Talwalkar.
\newblock Gradient descent on neural networks typically occurs at the edge of
  stability.
\newblock In \emph{International Conference on Learning Representations}, 2021.
\newblock URL \url{https://openreview.net/forum?id=jh-rTtvkGeM}.

\bibitem[Cubuk et~al.(2018)Cubuk, Zoph, Mane, Vasudevan, and
  Le]{cubuk2018autoaugment}
Ekin~D Cubuk, Barret Zoph, Dandelion Mane, Vijay Vasudevan, and Quoc~V Le.
\newblock Autoaugment: Learning augmentation policies from data.
\newblock \emph{arXiv preprint arXiv:1805.09501}, 2018.

\bibitem[Deng et~al.(2009)Deng, Dong, Socher, Li, Li, and
  Fei-Fei]{deng2009imagenet}
Jia Deng, Wei Dong, Richard Socher, Li-Jia Li, Kai Li, and Li~Fei-Fei.
\newblock Imagenet: A large-scale hierarchical image database.
\newblock In \emph{2009 IEEE conference on computer vision and pattern
  recognition}, pages 248--255. Ieee, 2009.

\bibitem[DeVries and Taylor(2017)]{devries2017improved}
Terrance DeVries and Graham~W Taylor.
\newblock Improved regularization of convolutional neural networks with cutout.
\newblock \emph{arXiv preprint arXiv:1708.04552}, 2017.

\bibitem[Donini et~al.(2020)Donini, Franceschi, Majumder, Pontil, and
  Frasconi]{donini2020marthe}
Michele Donini, Luca Franceschi, Orchid Majumder, Massimiliano Pontil, and
  Paolo Frasconi.
\newblock Marthe: Scheduling the learning rate via online hypergradients.
\newblock In Christian Bessiere, editor, \emph{Proceedings of the Twenty-Ninth
  International Joint Conference on Artificial Intelligence, {IJCAI-20}}, pages
  2119--2125. International Joint Conferences on Artificial Intelligence
  Organization, 7 2020.
\newblock \doi{10.24963/ijcai.2020/293}.
\newblock URL \url{https://doi.org/10.24963/ijcai.2020/293}.
\newblock Main track.

\bibitem[Dosovitskiy et~al.(2020)Dosovitskiy, Beyer, Kolesnikov, Weissenborn,
  Zhai, Unterthiner, Dehghani, Minderer, Heigold, Gelly,
  et~al.]{dosovitskiy2020image}
Alexey Dosovitskiy, Lucas Beyer, Alexander Kolesnikov, Dirk Weissenborn,
  Xiaohua Zhai, Thomas Unterthiner, Mostafa Dehghani, Matthias Minderer, Georg
  Heigold, Sylvain Gelly, et~al.
\newblock An image is worth 16x16 words: Transformers for image recognition at
  scale.
\newblock \emph{arXiv preprint arXiv:2010.11929}, 2020.

\bibitem[Gardner et~al.(2018)Gardner, Pleiss, Weinberger, Bindel, and
  Wilson]{gardner2018gpytorch}
Jacob Gardner, Geoff Pleiss, Kilian~Q Weinberger, David Bindel, and Andrew~G
  Wilson.
\newblock Gpytorch: Blackbox matrix-matrix gaussian process inference with gpu
  acceleration.
\newblock In S.~Bengio, H.~Wallach, H.~Larochelle, K.~Grauman, N.~Cesa-Bianchi,
  and R.~Garnett, editors, \emph{Advances in Neural Information Processing
  Systems}, volume~31. Curran Associates, Inc., 2018.
\newblock URL
  \url{https://proceedings.neurips.cc/paper/2018/file/27e8e17134dd7083b050476733207ea1-Paper.pdf}.

\bibitem[Ginsburg et~al.(2019)Ginsburg, Castonguay, Hrinchuk, Kuchaiev,
  Lavrukhin, Leary, Li, Nguyen, Zhang, and Cohen]{ginsburg2019novograd}
Boris Ginsburg, Patrice Castonguay, Oleksii Hrinchuk, Oleksii Kuchaiev, Vitaly
  Lavrukhin, Ryan Leary, Jason Li, Huyen Nguyen, Yang Zhang, and Jonathan~M
  Cohen.
\newblock Stochastic gradient methods with layer-wise adaptive moments for
  training of deep networks.
\newblock \emph{arXiv preprint arXiv:1905.11286}, 2019.

\bibitem[Goyal et~al.(2017)Goyal, Doll{\'a}r, Girshick, Noordhuis, Wesolowski,
  Kyrola, Tulloch, Jia, and He]{goyal2017gradual}
Priya Goyal, Piotr Doll{\'a}r, Ross Girshick, Pieter Noordhuis, Lukasz
  Wesolowski, Aapo Kyrola, Andrew Tulloch, Yangqing Jia, and Kaiming He.
\newblock Accurate, large minibatch sgd: Training imagenet in 1 hour.
\newblock \emph{arXiv preprint arXiv:1706.02677}, 2017.

\bibitem[He et~al.(2016)He, Zhang, Ren, and Sun]{he2016deep}
Kaiming He, Xiangyu Zhang, Shaoqing Ren, and Jian Sun.
\newblock Deep residual learning for image recognition.
\newblock In \emph{Proceedings of the IEEE conference on computer vision and
  pattern recognition}, pages 770--778, 2016.

\bibitem[Heo et~al.(2021)Heo, Chun, Oh, Han, Yun, Kim, Uh, and
  Ha]{heo2021adamp}
Byeongho Heo, Sanghyuk Chun, Seong~Joon Oh, Dongyoon Han, Sangdoo Yun, Gyuwan
  Kim, Youngjung Uh, and Jung-Woo Ha.
\newblock Adamp: Slowing down the slowdown for momentum optimizers on
  scale-invariant weights.
\newblock In \emph{International Conference on Learning Representations}, 2021.
\newblock URL \url{https://openreview.net/forum?id=Iz3zU3M316D}.

\bibitem[Hoffer et~al.(2017)Hoffer, Hubara, and Soudry]{hoffer2017longer}
Elad Hoffer, Itay Hubara, and Daniel Soudry.
\newblock Train longer, generalize better: closing the generalization gap in
  large batch training of neural networks.
\newblock In I.~Guyon, U.~V. Luxburg, S.~Bengio, H.~Wallach, R.~Fergus,
  S.~Vishwanathan, and R.~Garnett, editors, \emph{Advances in Neural
  Information Processing Systems}, volume~30, pages 1731--1741. Curran
  Associates, Inc., 2017.

\bibitem[Kingma and Ba(2014)]{kingma2014adam}
Diederik~P Kingma and Jimmy Ba.
\newblock Adam: A method for stochastic optimization.
\newblock \emph{arXiv preprint arXiv:1412.6980}, 2014.

\bibitem[Krizhevsky(2014)]{krizhevsky2014weird}
Alex Krizhevsky.
\newblock One weird trick for parallelizing convolutional neural networks.
\newblock \emph{arXiv preprint arXiv:1404.5997}, 2014.

\bibitem[Krizhevsky and Hinton(2009)]{krizhevsky2009learning}
Alex Krizhevsky and Geoffrey Hinton.
\newblock Learning multiple layers of features from tiny images.
\newblock 2009.

\bibitem[Loshchilov and Hutter(2017)]{loshchilov2016sgdr}
Ilya Loshchilov and Frank Hutter.
\newblock Sgdr: Stochastic gradient descent with warm restarts.
\newblock 2017.

\bibitem[M{\"u}ller et~al.(2019)M{\"u}ller, Kornblith, and
  Hinton]{muller2019does}
Rafael M{\"u}ller, Simon Kornblith, and Geoffrey Hinton.
\newblock When does label smoothing help?
\newblock \emph{arXiv preprint arXiv:1906.02629}, 2019.

\bibitem[Picheny et~al.(2020)Picheny, Dutordoir, Artemev, and
  Durrande]{victor2020bo}
Victor Picheny, Vincent Dutordoir, Artem Artemev, and Nicolas Durrande.
\newblock Automatic tuning of stochastic gradient descent with bayesian
  optimisation.
\newblock \emph{arXiv preprint arXiv:2006.14376}, 2020.

\bibitem[Rasmussen et~al.(2006)Rasmussen, Williams, Press, Bach, and
  (Firm)]{rasmussen2006gaussian}
C.E. Rasmussen, C.K.I. Williams, M.I.T. Press, F.~Bach, and ProQuest (Firm).
\newblock \emph{Gaussian Processes for Machine Learning}.
\newblock Adaptive computation and machine learning. MIT Press, 2006.
\newblock ISBN 9780262182539.

\bibitem[{Schraudolph}(1999)]{schraudolph1999local}
N.~N. {Schraudolph}.
\newblock Local gain adaptation in stochastic gradient descent.
\newblock In \emph{1999 Ninth International Conference on Artificial Neural
  Networks ICANN 99. (Conf. Publ. No. 470)}, volume~2, pages 569--574 vol.2,
  1999.
\newblock \doi{10.1049/cp:19991170}.

\bibitem[Smith and Topin(2019)]{smith2019super}
Leslie~N Smith and Nicholay Topin.
\newblock Super-convergence: Very fast training of neural networks using large
  learning rates.
\newblock In \emph{Artificial Intelligence and Machine Learning for
  Multi-Domain Operations Applications}, volume 11006, page 1100612.
  International Society for Optics and Photonics, 2019.

\bibitem[Tan and Le(2019)]{tan2019efficientnet}
Mingxing Tan and Quoc Le.
\newblock Efficientnet: Rethinking model scaling for convolutional neural
  networks.
\newblock In \emph{International Conference on Machine Learning}, pages
  6105--6114. PMLR, 2019.

\bibitem[Touvron et~al.(2020)Touvron, Cord, Douze, Massa, Sablayrolles, and
  J{\'e}gou]{touvron2020deit}
Hugo Touvron, Matthieu Cord, Matthijs Douze, Francisco Massa, Alexandre
  Sablayrolles, and Herv{\'e} J{\'e}gou.
\newblock Training data-efficient image transformers \& distillation through
  attention.
\newblock \emph{arXiv preprint arXiv:2012.12877}, 2020.

\bibitem[You et~al.(2017)You, Gitman, and Ginsburg]{you2017lars}
Yang You, Igor Gitman, and Boris Ginsburg.
\newblock Large batch training of convolutional networks.
\newblock \emph{arXiv preprint arXiv:1708.03888}, 2017.

\bibitem[You et~al.(2020)You, Li, Reddi, Hseu, Kumar, Bhojanapalli, Song,
  Demmel, Keutzer, and Hsieh]{you2020lamb}
Yang You, Jing Li, Sashank Reddi, Jonathan Hseu, Sanjiv Kumar, Srinadh
  Bhojanapalli, Xiaodan Song, James Demmel, Kurt Keutzer, and Cho-Jui Hsieh.
\newblock Large batch optimization for deep learning: Training bert in 76
  minutes.
\newblock In \emph{International Conference on Learning Representations}, 2020.
\newblock URL \url{https://openreview.net/forum?id=Syx4wnEtvH}.

\bibitem[Zagoruyko and Komodakis(2016)]{zagoruyko2016wide}
Sergey Zagoruyko and Nikos Komodakis.
\newblock Wide residual networks.
\newblock \emph{arXiv preprint arXiv:1605.07146}, 2016.

\bibitem[Zhang et~al.(2020)Zhang, Lang, Liu, and Xiao]{zhang2020salsa}
Pengchuan Zhang, Hunter Lang, Qiang Liu, and Lin Xiao.
\newblock Statistical adaptive stochastic gradient methods.
\newblock \emph{arXiv preprint arXiv:2002.10597}, 2020.

\end{thebibliography}


\newpage

\appendix
\section{More Details}
\label{app:details}
The overall algorithm of \autowu\ is depicted in Algorithm \ref{alg:main}.
All implementations are based on PyTorch 1.7.0 and GPyTorch 1.3.0. Experiments are conducted on NVIDIA Tesla V100 GPUs with 32GB memory.

We use the weight decay of $0.1$, $\beta_1 = 0.9$, $\beta_2 = 0.999$, $\epsilon = 10^{-8}$ and $\delta = 0.1$ for AdamP and the weight decay of $0.1$, $\beta_1 = 0.9$, $\beta_2 = 0.999$ and $\epsilon = 10^{-6}$ for LAMB.

In all experiments involving CIFAR-10 or CIFAR-100, the corresponding network is trained with AutoAugment \citep{cubuk2018autoaugment}, Cutout \citep{devries2017improved}, and label smoothing \citep{muller2019does} with factor 0.1 for 200 epochs. We report the mean and the standard deviation from three independent runs with different random seeds. In case of ResNet-50 training on ImageNet, the standard set of augmentations as in \citet{he2016deep} is used for ImageNet and the network is trained for 120 epochs. We report the result of a single run with a fixed random seed in all ImageNet experiments due to resource constraints.

\begin{algorithm}[tb]
\small
  \caption{Automated LR scheduler (\autowu)}
  \label{alg:main}
    \begin{algorithmic}
        \State {\bfseries Hyperparameters:} $\eta_{\min}$, $\eta_{\max}$, $\rho_w \in (0, 1)$, $n_{test}$, $c$ (confidence), $p$ (patience)
        \State {\bfseries Inputs:} $\theta_0$, $\mathcal{A}$ (optimizer), $T$ (the number of total steps)
        \State $\eta_0 \gets \eta_{\min}$, $\gamma \gets (\eta_{\max}/\eta_{\min})^{1/\lfloor \rho_w T\rfloor}$, 
        $\mathrm{patience\_flag} \gets 0$, $\mathcal{C} \gets \emptyset$
        \For{$t$ {\bfseries from} $0$ {\bfseries to} $T-1$}
            \State Sample $\xi_t$. 
            \State Compute loss and stochastic gradient: $L_t = L(\theta_t; \xi_t)$, $g_t = \nabla_{\theta_t} L_t$.
            \State Take an optimizer step of $\mathcal{A}$ w.r.t. LR $\eta_t$ and $g_t$ to compute $\theta_{t+1}$.
            \If{warmup phase}
                \State $\mathcal{C} \gets \mathcal{C} \cup \{(t, L_t)\}$.
                \If{end of epoch}
                    \State Subsample $\mathcal{C}_0, \cdots, \mathcal{C}_{n_{test}}$ from $\mathcal{C}$.
                    \State Fit GP parameters w.r.t. marginal log-likelihood of $\mathcal{C}_0$.
                    \State Infer $f_i$ conditioned on $\mathcal{C}_i$ for $i=1,\cdots,n_{test}$.
                    \State Compute $P_{\min}(f_i)$ as in Eqn. (\ref{eqn:lb_prob}).
                    \If{$\#(i: P_{\min}(f_i) > c) > n_{test}/2$}
                        \State $\mathrm{patience\_flag} \gets \mathrm{patience\_flag} + 1$
                    \Else
                        \State $\mathrm{patience\_flag} \gets 0$
                    \EndIf
                    \If{$\mathrm{patience\_flag} \geq p$}
                        \State Compute $t^*$ according to Eqn. (\ref{eqn:tstar}).
                        \State Set $\eta_{t+1} \gets \gamma^{t^*} \eta_0$ and switch to decay phase.
                    \EndIf
                \EndIf
                \State Update LR if not switched: $\eta_{t+1} \gets \gamma \eta_t$.
            \Else
                \State Compute $\eta_{t+1}$ according to cosine or cosine-then-decay schedule.
            \EndIf
        \EndFor
    \end{algorithmic}
\end{algorithm}

\section{Further Evaluations}
\label{app:further_eval}

In this section, further evaluation results which were not discussed in Section \ref{sec:experiments} are presented. 

\begin{figure}[t]
    \centering
    \begin{subfigure}[b]{0.3\textwidth}
    \includegraphics[width=\textwidth]{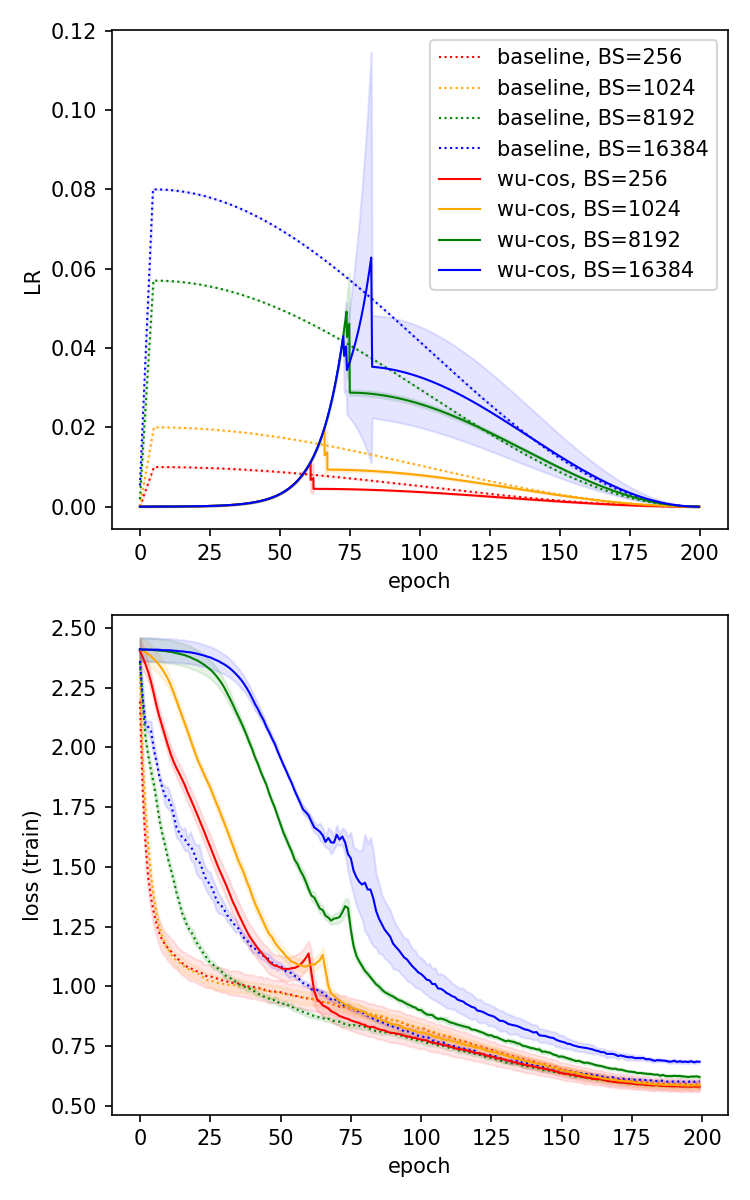}
    \caption{CIFAR-10}
    \end{subfigure}
    \hspace{10mm}
    \begin{subfigure}[b]{0.3\textwidth}
    \includegraphics[width=\textwidth]{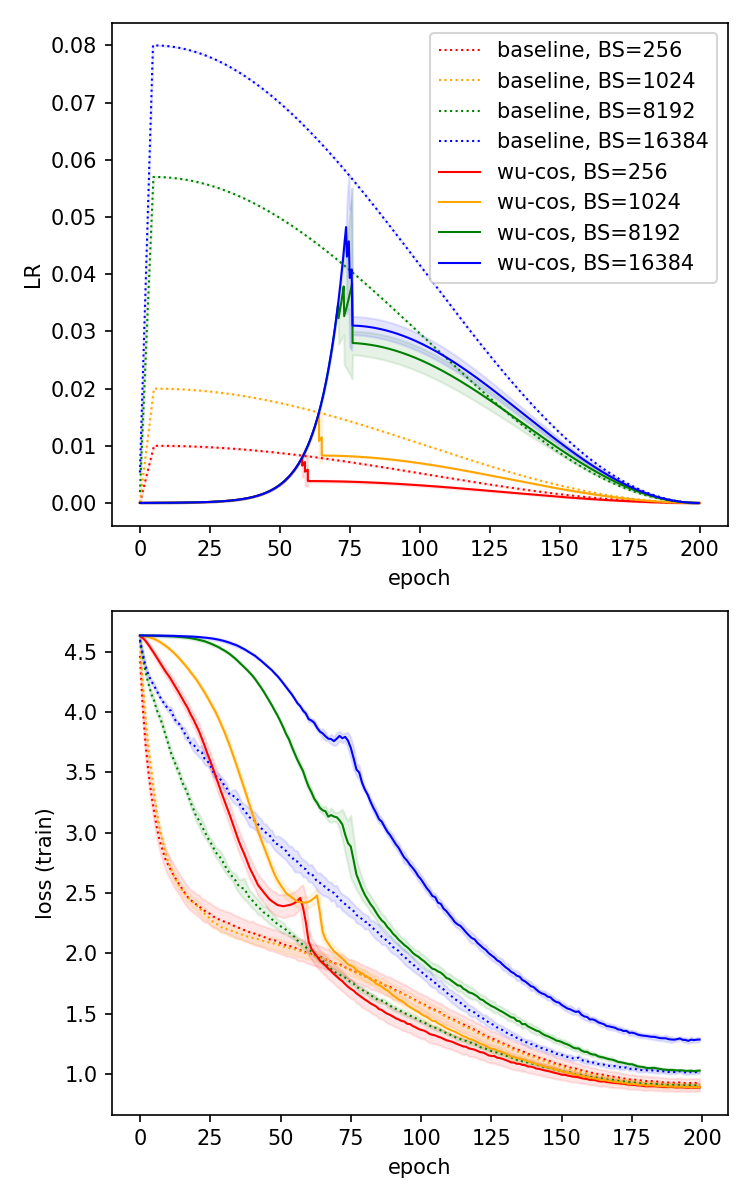}
    \caption{CIFAR-100}
    \end{subfigure}
    \caption{LR schedules (top row) and training loss curves (bottom row) of the baseline and \autowu\ with cosine decay on CIFAR tasks, when the base optimizer is LAMB.}
    \label{fig:lamb_exp}
\end{figure}

Firstly, the comparison between \autowu\ and the baseline schedule when coupled with LAMB is made. The peak LR of the baseline is scaled as $\eta_{base} \sqrt{B/256}$ for batch size $B$. The base LR $\eta_{base}$ is chosen $0.01$ rather than $0.001$, since the latter has resulted in worse performances. The results are summarized in Table \ref{tab:exp_cifar}, and the plot of LR schedules and training loss curves can be found in Figure \ref{fig:lamb_exp}. We observe that general tendency is similar to that of \autowu\ coupled with AdamP, but the resulting performances are slightly worse than those obtained by the baseline schedule with an exception of batch size 8192 on CIFAR-100. However, we emphasize that \autowu\ does not require tuning, in contrast to the fact that LAMB can be more sensitive to $\eta_{base}$ \citep{xinlei20201mocov3} and adjusting $\eta_{base}$ to 0.01 in baselines was necessary to achieve the reported performance.

\begin{table*}[t]
\scriptsize
    \centering
    \caption{Comparison of test accuracies (\%) on CIFAR-10 and CIFAR-100 between a default schedule and \autowu\ with LAMB.}
    \begin{tabular}{cc|cccc}
    \hline
    \multirow{2}{*}{\makecell{Dataset \\ (Architecture)}}
    & \multirow{2}{*}{Schedule} 
    & \multicolumn{4}{c}{Batch size} 
    \\
    & & 256 & 1K & 8K & 16K
    \\
    \hline
    \multirow{3}{*}{\makecell{CIFAR-10 \\ (ResNet-18)}} 
    & Baseline
    & \textbf{96.37 {\tiny (0.16)}}
    & \textbf{96.39 {\tiny (0.14)}}
    & \textbf{95.92 {\tiny (0.09)}}
    & \textbf{95.36 {\tiny (0.08)}}
    \\
    & \autowu\ + const-cos
    & 96.26 {\tiny (0.09)}
    & 96.17 {\tiny (0.07)}
    & 95.53 {\tiny (0.13)}
    & 94.47 {\tiny (0.23)}
    \\
    & \autowu\ + cos
    & 96.23 {\tiny (0.14)}
    & 96.14 {\tiny (0.09)}
    & 95.37 {\tiny (0.10)}
    & 94.19 {\tiny (0.22)}
    \\
    \hline
    \multirow{3}{*}{\makecell{CIFAR-100 \\ (WideResNet28-10)}} 
    & Baseline
    & \textbf{83.38 {\tiny (0.15)}} 
    & \textbf{83.61 {\tiny (0.01)}} 
    & 82.16 {\tiny (0.19)}
    & \textbf{79.99 {\tiny (0.30)}}
    \\
    & \autowu\ + const-cos
    & 82.86 {\tiny (0.16)}
    & 83.11 {\tiny (0.32)} 
    & \textbf{82.44 {\tiny (0.11)}}
    & 79.36 {\tiny (0.47)}
    \\
    & \autowu\ + cos
    & 82.92 {\tiny (0.33)}
    & 82.88 {\tiny (0.38)} 
    & 81.31 {\tiny (0.06)}
    & 77.57 {\tiny (0.50)}
    \\  
    \hline
    \end{tabular} 
    \label{tab:exp_cifar}
\normalsize
\end{table*}

\begin{figure}[t]
    \centering
    \begin{subfigure}[b]{0.3\textwidth}
    \includegraphics[width=\textwidth]{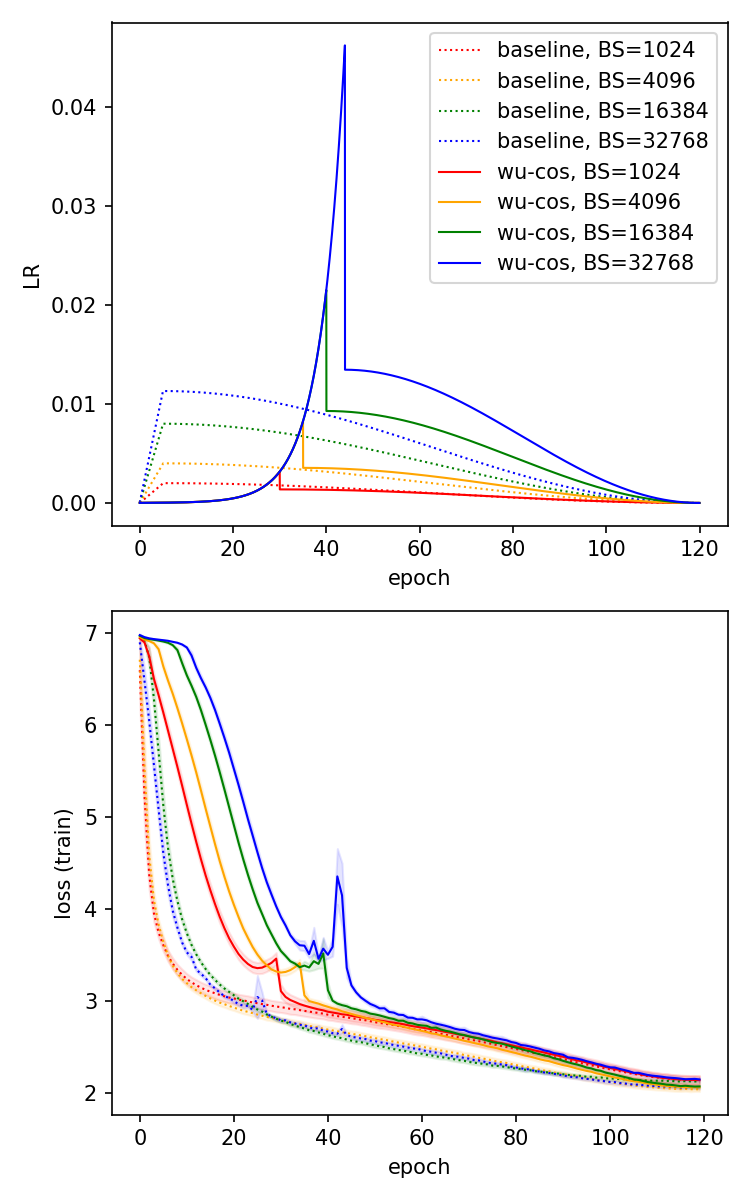}
    \caption{EfficientNet-B0}
    \end{subfigure}
    \hspace{10mm}
    \begin{subfigure}[b]{0.3\textwidth}
    \includegraphics[width=\textwidth]{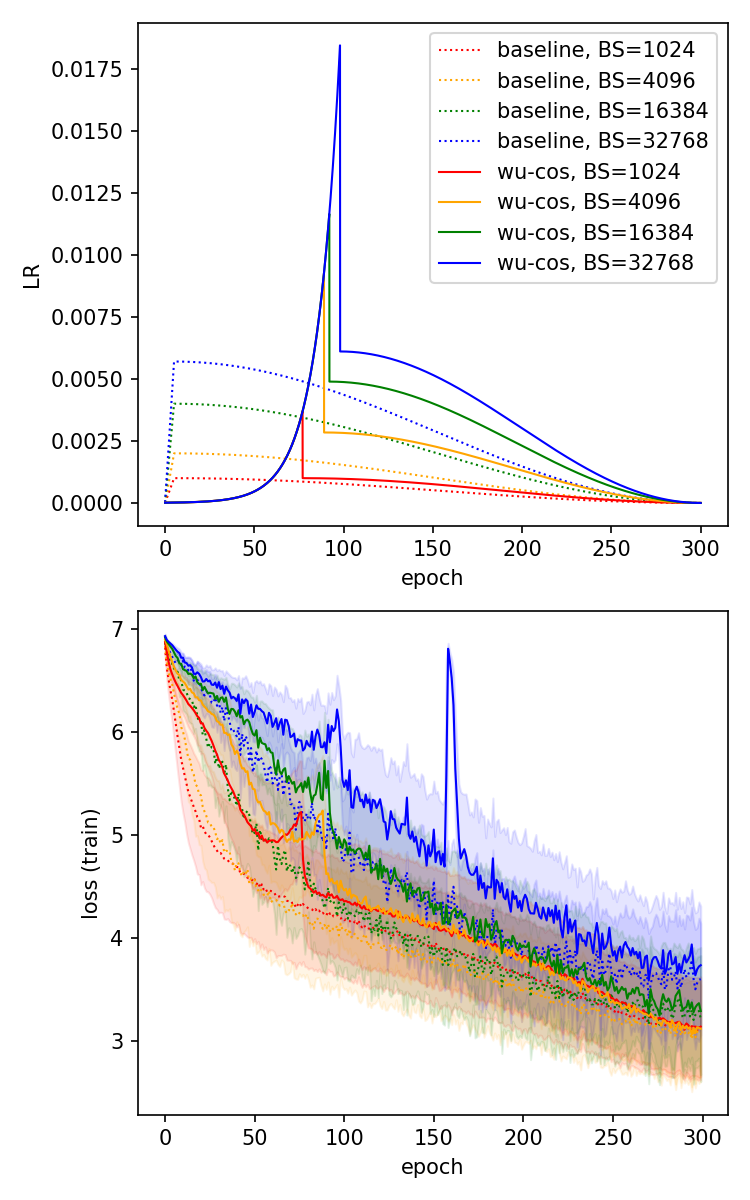}
    \caption{ViT-S/16}
    \end{subfigure}
    \caption{LR schedules (top row) and training loss curves (bottom row) of the baseline and \autowu\ with cosine decay on ImageNet with EfficientNet-B0 and ViT-S/16, when the base optimizer is AdamP.}
    \label{fig:exp_effnet_vit}
\end{figure}

We also evaluate the performance of \autowu\ on ImageNet classification task with two other architectures: EfficientNet-B0 and ViT-S/16. The training configuration of EfficientNet-B0 is identical to that of ResNet-50, except that the label smoothing of factor 0.1 is used. For ViT-S/16, it is trained for 300 epochs with augmentations as described in \citet{touvron2020deit}. The peak LR of the baseline is scaled as $\eta_{base} \sqrt{B/256}$ with $\eta_{base}=0.001$ for EfficientNet-B0 and $\eta_{base}=0.0005$ for ViT-S/16. Results are found in Table \ref{tab:exp_effnet_vit}, and the plots of LR schedules and training loss curves are found in Figure \ref{fig:exp_effnet_vit}. Even though the performances are a little bit lower compared to the tuned baseline, \autowu\ stably works on other architectures without any specific hyperparameter tuning.

\begin{table*}[t]
\small
    \centering
    \caption{Comparison of top-1 validation accuracies (\%) between the baseline schedule and \autowu, in case of ImageNet training on EfficientNet-B0 and ViT-S/16 with AdamP.}
    \begin{tabular}{cc|cccc}
    \hline
    \multirow{2}{*}{\makecell{Dataset \\ (Architecture)}} 
    & \multirow{2}{*}{Schedule} 
    & \multicolumn{4}{c}{Batch size} \\
    & & 1K & 4K & 16K & 32K \\
    \hline
    \multirow{3}{*}{\makecell{ImageNet \\ (EfficientNet-B0)}} 
    & Baseline 
    & \textbf{75.03} & \textbf{76.00} & 74.23 & \textbf{75.17}
    \\
    & \autowu\ + const-cos
    & 74.58 & 75.43 & 75.01 & 73.99 
    \\
    & \autowu\ + cos
    & 74.90 & 75.81 & \textbf{75.44} & 74.34 
    \\
    \hline
    \multirow{3}{*}{\makecell{ImageNet \\ (ViT-S/16)}} 
    & Baseline 
    & \textbf{79.37} & \textbf{79.34} & \textbf{77.39} & \textbf{73.92}
    \\
    & \autowu\ + const-cos
    & 77.65 & 78.14 & 75.67 & 72.73 
    \\
    & \autowu\ + cos
    & 79.01 & 79.15 & 76.95 & 72.38 
    \\
    \hline
    \end{tabular} 
    \label{tab:exp_effnet_vit}
\normalsize
\end{table*}

\section{Ablation Studies}
\label{app:abl_studies}

\subsection{Sensitivity of the baseline with respect to the warmup schedule}
\label{app:abl_grid_search}

We compare the performances of the baseline LR scheduling according to various numbers of warmup epochs and peak LRs. Namely, the LR is linearly increased from 0 to the peak LR $\eta_{peak}$ for $n_{warmup}$ epochs then decayed to 0 via cosine schedule, and the experiment is carried out for $\eta_{peak} \in \{0.002, 0.004, 0.008, 0.016, 0.032\}$ and $n_{warmup} \in \{5, 20, 40, 60\}$. Here, we use ResNet-50 trained by AdamP with a batch size of 16384 on ImageNet, and therefore the configuration $(\eta_{peak}, n_{warmup}) = (0.008, 5)$ corresponds to the baseline schedule (presented in Table \ref{tab:main_exp}).

\begin{figure}[t]
    \centering
    \includegraphics[width=0.65\textwidth]{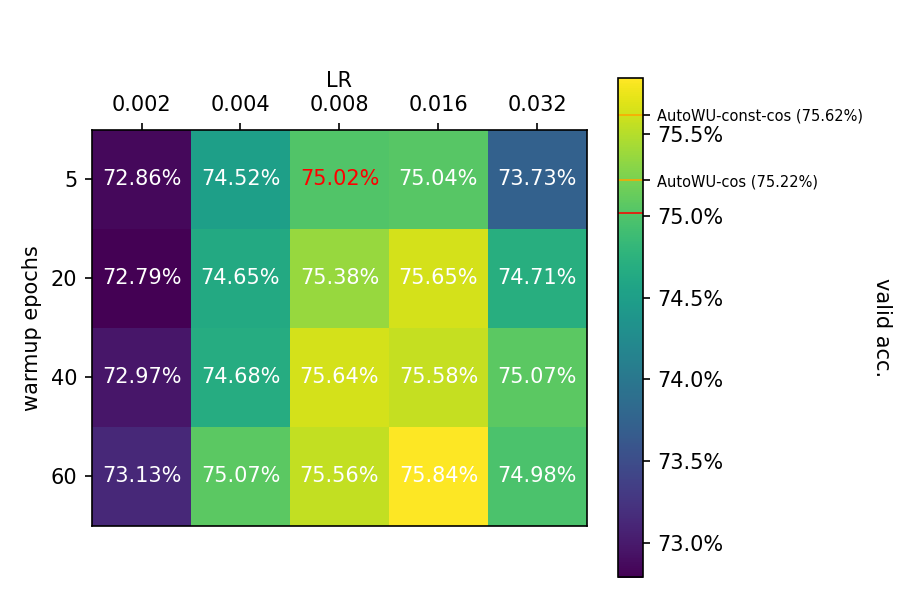}
    \caption{Comparison of validation accuracies over the choices to the peak LR $\eta_{peak}$ and the number of warmup epochs $n_{warmup}$, in case of ResNet-50 training on ImageNet with AdamP and a batch size of 16384. The schedule presented as the baseline in Table \ref{tab:main_exp} corresponds to $(\eta_{peak}, n_{warmup})=(0.008, 5)$ and colored red in the plot. Additionally, the performances of \autowu\ with cosine and constant-then-cosine decay are annotated as orange lines in the colorbar on the right.}
    \label{fig:abl_grid_search}
\end{figure}

Figure \ref{fig:abl_grid_search} demonstrates how the performance of ResNet-50 on ImageNet changes with respect to $\eta_{peak}$ and $n_{warmup}$. The value of optimal LR is either 0.008 or 0.016 when $n_{warmup}$ is fixed, which includes our baseline. On the other hand, we find that the validation accuracy is improved as $n_{warmup}$ increases and the best performance is 75.84\% obtained by $(\eta_{peak}, n_{warmup}) = (0.016, 60)$. We remark that not only the performances of \autowu\ with cosine and constant-then-cosine decay (75.22\% and 75.62\%, respectively) are fairly close to the best performance but also the peak LR and the warmup epochs found by \autowu , $(0.0215, 40)$ as shown in Figure \ref{fig:main_exp}(c), are similar to the best configuration on the baseline schedule. This validates the effectiveness of the proposed algorithm.

\subsection{Dependency of \autowu\ on the warmup schedule}
\label{app:abl_warmup}

We have argued that the exponential growth in the warmup phase stabilizes the training and enables a fine-grained LR exploration. To demonstrate this, we compare the linear growth and the exponential growth in case of ResNet-50 training on ImageNet with AdamP and batch size 16384. Specifically, we consider the linear schedule in the warmup phase defined as
\begin{equation}
   \eta_t = \eta_{\min} + (\eta_{\max} - \eta_{\min}) \cdot \frac{t}{\lfloor \rho_{w} T \rfloor} 
   \quad \text{for} \quad 
   t \in \{0, \cdots, \lfloor \rho_w T \rfloor\},
\end{equation}
where $\eta_{\min} = 10^{-5}$ and $\eta_{\max} \in \{0.1, 1.0\}$. In both choices of $\eta_{\max}$, the cosine schedule is used in the decay phase and all hyperparameters are set identical as in case of \autowu\ with exponential warmup schedule. 

\begin{figure}[t]
    \centering
    \includegraphics[width=0.9\textwidth]{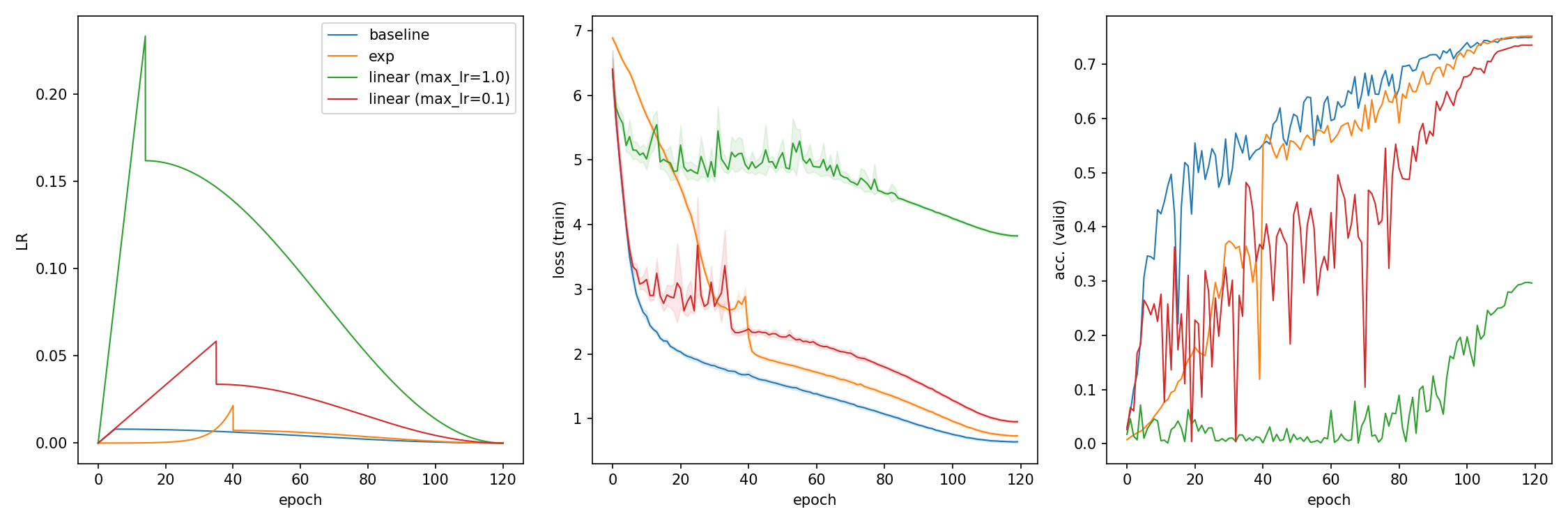}
    \caption{Comparison of the linear growth and the exponential growth in the warmup phase of \autowu. ResNet-50 is trained on ImageNet with a batch size of 16384, AdamP optimizer, and the corresponding scheduler.}
    \label{fig:linear_warmup}
\end{figure}

Figure \ref{fig:linear_warmup} shows the LR schedule, the training loss and the validation accuracy of the baseline, \autowu\ with exponential growth, and \autowu\ with linear growth as described above. In terms of the validation accuracy, the baseline and \autowu\ with exponential growth achieves 75.02\% and 75.22\% respectively, while \autowu\ with linear growth achieves 29.65\% for $\eta_{\max} = 1.0$ and 73.56\% for $\eta_{\max} = 0.1$. This implies that the linear warmup makes \autowu\ very sensitive to the choice of $\eta_{\max}$, hence a significant amount of effort to tune $\eta_{\max}$ must be made, obliterating the purpose of automated LR scheduling. 

\begin{table}[t]
\small
    \centering
    \caption{Sensitivity of \autowu\ (+cos) with respect to the maximum fraction of warmup ($\rho_w$).}
    \begin{tabular}{c|cccc}
    \hline\hline
    \multirow{2}{*}{$\rho_w$} & 
    \multicolumn{4}{c}{Batch size} \\
    & 1K & 4K & 16K & 32K \\
    \hline
    0.125 
    & 75.89 & 75.59 & 74.52 & 73.40
    \\
    0.25 
    & \textbf{76.60} & \textbf{76.04} & \textbf{75.28} & 73.89
    \\
    0.5
    & 76.19 & 75.70 & 75.22 & \textbf{74.40}
    \\
    \hline\hline
    \end{tabular} 
    \label{tab:exp_maxwu}
\normalsize
\end{table}
The growth factor $\gamma$ in Eqn. \ref{eqn:warmup_sched} in \autowu\ with the exponential growth is completely determined by $\rho_w$ when $\eta_{\min}$ and $\eta_{\max}$ are set to the sufficiently small and large values; $\gamma$ is increased if $\rho_w$ is decreased. Therefore, we also compare the performances of \autowu\ with the cosine decay for different values of $\rho_w$.

Specifically, ResNet-50 is trained on ImageNet with a batch size of 16384, AdamP, and \autowu\ for $\rho_w \in \{0.125, 0.25, 0.5\}$, and the results are summarized in Table \ref{tab:exp_maxwu}. Here, $\rho_w = 0.5$ corresponds to the default configuration which is also reported in Table \ref{tab:main_exp}. The best performances are attained by $\rho_w=0.25$ or $\rho_w = 0.5$, and when $\rho_w = 0.125$, the performances are degraded.

\begin{figure}[t]
    \centering
    \includegraphics[width=0.5\textwidth]{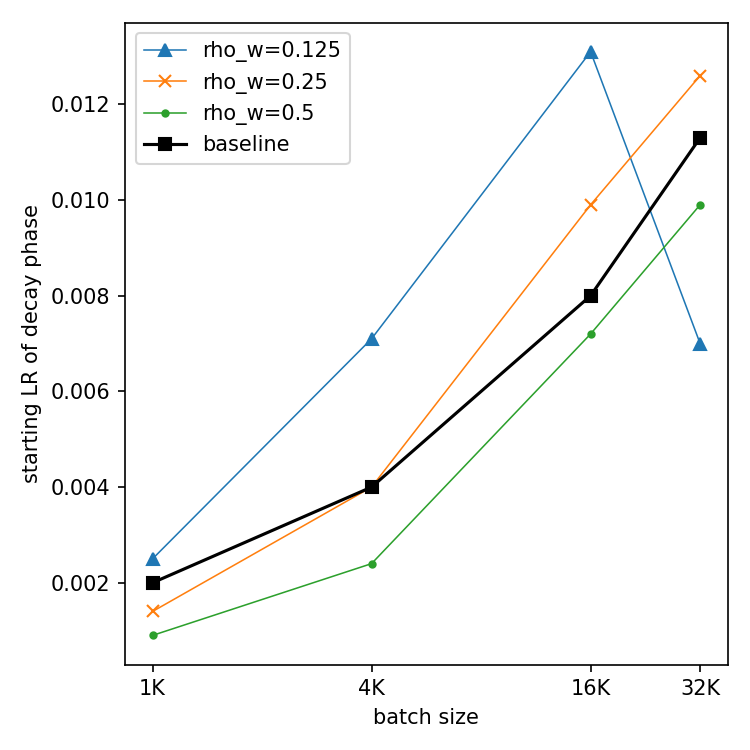}
    \caption{Dependency of \autowu\ on $\rho_w$ in terms of the starting LR of the decay phase. Plotted are the starting LRs of the decay phase of \autowu\ for $\rho_w \in \{0.125, 0.25, 0.5\}$ together with the peak LRs of the baseline, for batch sizes $\{1024, 4096, 16384, 32768\}$.}
    \label{fig:abl_maxwu_lr}
\end{figure}

We plot the relation between the starting LR $\eta_{t^*}$ of the decay phase and the choice of $\rho_w$ in Figure \ref{fig:abl_maxwu_lr}. Larger batch size or faster growth implies the bigger starting LR in the decay phase in general, but this breaks down when the batch size becomes very large. We have observed that this ``critical batch size'' is larger with larger $\rho_w$ (\emph{i.e.} smaller $\gamma$), and supports the intuition that slow growth of LR in the warmup phase stabilizes the overall training dynamics. Such an observation is also consistent with the previous works \citep{smith2019super, cohen2021gradient}.

\newpage
\section{Further discussions}

We did not present the study investigating the impact of other hyperparameters including the confidence $c$, patience $p$, and the choice of decay schedule on the performance of the proposed algorithm \autowu. While we leave a thorough study as a future work, we briefly discuss below on a few observations from our preliminary experiments.
\begin{itemize}
    \item It seems that the choice of confidence $c$ does not matter much, as $P_{\min}(f)$ in Eqn. (\ref{eqn:lb_prob}) become very close to 1 when the loss starts to increase. We also note that conducting $n_{test} > 1$ tests on sub-sampled loss curves made the test more robust. 
    \item The choice of $p$ was more critical. Note that \autowu\ is an algorithm which detects whether the loss minimum lies in the past or not; this implies that \autowu\ may be susceptible to a local fluctuation of loss, especially when the number of steps is small (or equivalently, the batch size is large). To reduce such a sensitivity, the patience $p$ to wait whether the observed minimum is local or not is introduced. Utilizing a large value of $p$ makes the test more robust, but simultaneously lets LR to grow until a large value hence training may diverge. We found that setting $p=3$ was working well for all experiments. 
    
    \item We argued that SALSA \citep{zhang2020salsa} is the most similar work to \autowu. In terms of warmup scheduling, SALSA employs a line search based method to warmup the initial training dynamics and ends the warmup phase \emph{when the loss stops to decrease}, in the similar spirit as this work. However, the extra computation required for the line search was less appealing than a simple exponential schedule, and we have observed that the line search based method in general is sensitive to the Armijo constant especially for large batch sizes.
\end{itemize}

\end{document}